\documentclass[letterpaper, 10pt, conference]{ieeeconf}
\usepackage{amsmath,amsfonts}
\usepackage{algorithmic}
\usepackage{algorithm}
\usepackage{array}
\usepackage[caption=false,font=normalsize,labelfont=sf,textfont=sf]{subfig}
\usepackage{textcomp}
\usepackage{stfloats}
\usepackage{url}
\usepackage{verbatim}
\usepackage{graphicx}
\usepackage{cite}
\usepackage[hidelinks]{hyperref}
\usepackage{cleveref}
\usepackage[hidelinks]{hyperref}
\usepackage{comment}
\begin{document}

\title{Proprioceptive Shape Estimation of Tensegrity \\Manipulators Using Energy Minimisation
}

\author{
    Tufail Ahmad Bhat$^{1}$,
    Shuhei Ikemoto$^{1}$
\thanks{This work was supported by JSPS KAKENHI Grant Number 25H02619.}
\thanks{$^{1}$ Both authors are with the Graduate School of Life Science and Systems Engineering, Kyushu Institute of Technology, 2-4 Hibikino, Wakamatsu, Kitakyushu, Fukuoka, Japan.n (e-mail:  {\tt\small ikemoto@brain.kyutech.ac.jp})  }
}
\IEEEoverridecommandlockouts

\maketitle
\vspace{-2mm}
\begin{abstract}
Shape estimation is fundamental for controlling continuously bending tensegrity manipulators, yet achieving it remains a challenge. Although using exteroceptive sensors makes the implementation straightforward, it is costly and limited to specific environments. Proprioceptive approaches, by contrast, do not suffer from these limitations. So far, several methods have been proposed; however, to our knowledge, there are no proven examples of large-scale tensegrity structures used as manipulators. This paper demonstrates that shape estimation of the entire tensegrity manipulator can be achieved using only the inclination angle information relative to gravity for each strut. Inclination angle information is intrinsic sensory data that can be obtained simply by attaching an inertial measurement unit (IMU) to each strut. Experiments conducted on a five-layer tensegrity manipulator with 20 struts and a total length of 1160 mm demonstrate that the proposed method can estimate the shape with an accuracy of 2.1 \% of the total manipulator length, from arbitrary initial conditions under both static conditions and maintains stable shape estimation under external disturbances.                                                                                                                                                                                                                                                                                                                                                                                                                                                                                                                                                                                                                                                                                                                                                                                                                                               

%Shape estimation of deformable manipulators is fundamental for their control. Exteroceptive sensor-based shape estimation is relatively easy to implement but cost-intensive, and not feasible for open and non-continuous lattice structures like tensegrity manipulators, due to occlusions. For relatively simple structures, both exteroceptive and proprioceptive sensor-based methods have been explored. However, to the best of our knowledge, none of them have been demonstrated on a full-scale tensegrity manipulator. Here, we present the first proprioceptive shape estimator for a tensegrity manipulator, using Inertial Measurement Unit (IMU) sensors mounted on the strut elements. Our method uses inclination angles of the strut elements and computes the spatial nodal positions using energy minimisation. We test the method on a 1160 mm long custom-made Five-layer tensegrity manipulator. The experiments demonstrated that the proposed method can estimate the shape within 2.1 \% of the tensegrity manipulator length in the static case and in the presence of external disturbances. This highly redundant flexible manipulator with self-shape awareness could enable safer and more intelligent interactions in unknown environments.

\end{abstract}

\section{Introduction}

\begin{figure}[t]
    \centering
    \includegraphics[width=\linewidth]{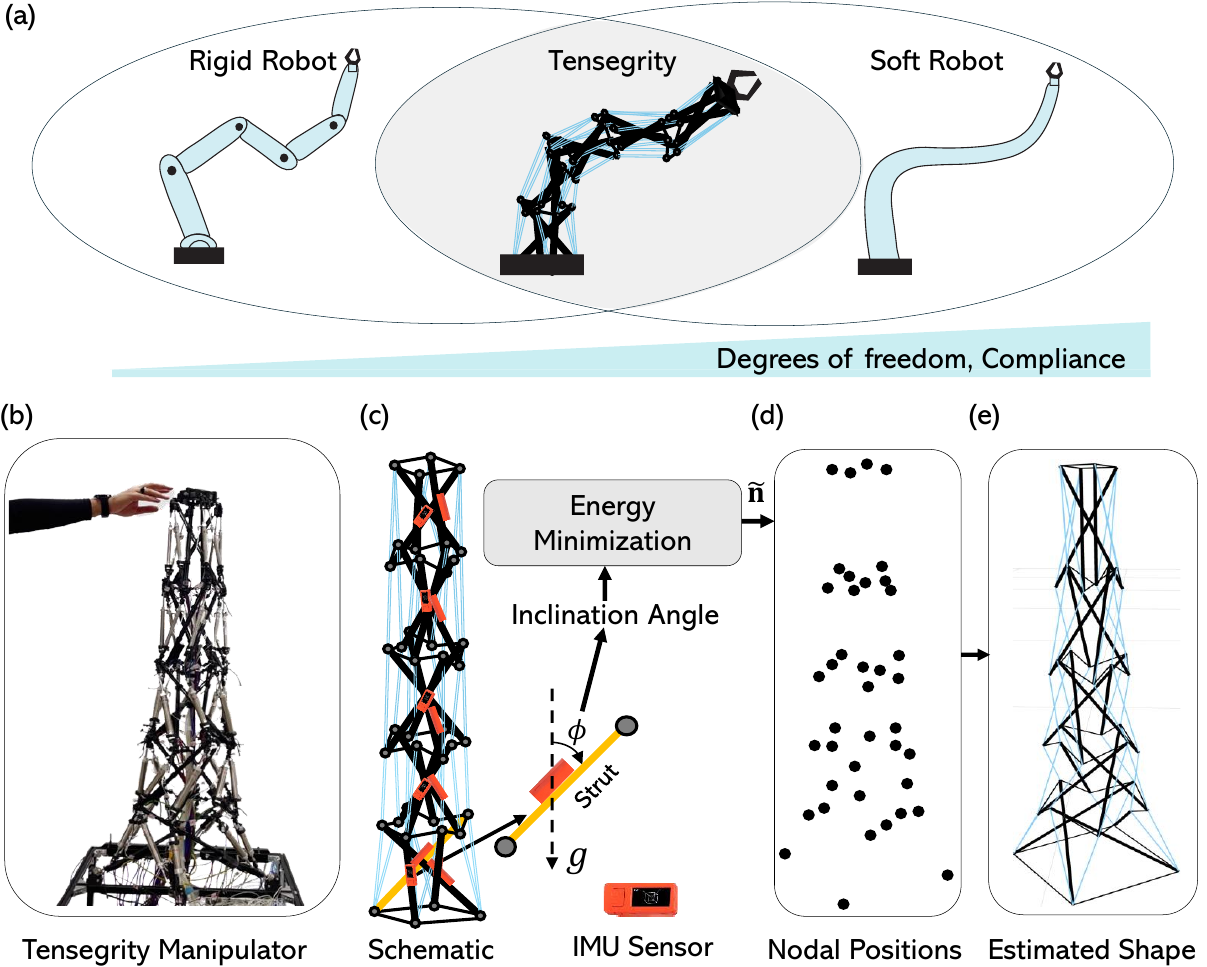}

    \caption{
   Tensegrity manipulator with self-shape awareness
    a) The tensegrity manipulator provides a balance between rigidity and flexibility. 
    b) \textbf{ TM40}: a highly redundant five-layer tensegrity manipulator. c) Schematic of the three-dimensional TM40 tensegrity structure with the proprioceptive sensors, used for the energy minimisation algorithm. d) Estimated nodal positions. e) Full-scale estimated shape of the tensegrity manipulator determined from the spatial nodal positions.
    }
    \label{fig:TM40 tensegrity manipulator}
 
\end{figure}

Biological systems often inspire advances in robotics. For example, continuum robots draw inspiration from natural structures such as octopus tentacles \cite{xie2023octopus} or elephant trunks \cite{peng2025dexterous}. In theory, these robots possess an infinite number of degrees of freedom (DOF), and offer high compliance and adaptability. This allows them to interact safely with humans in collaborative environments \cite{cangan2022model}, where traditional rigid manipulators often fail. However, these very properties make them hard to model and control \cite{liu2022morphology, huang2024predicting}. In practice, only a limited number of DOF can be controlled using their actuators, which limits their controllability and precision \cite{do2024stiffness}. Furthermore, their load carrying capacity is often limited. An alternative design that addresses some of these limitations is a tensegrity manipulator, which consists of rigid links, known as struts, and flexible elements, such as cables \cite{skelton2009tensegrity}. It provides a tunable trade-off between rigidity and flexibility, illustrated in Fig.~\ref{fig:TM40 tensegrity manipulator} (a). Tensegrity-based manipulators \cite{ramadoss2022hedra, lessard2016bio, fadeyev2019generalized,kobayashi2023active,10669185,kobayashi2022soft, 11246578,hsieh2024adaptive } are actively being studied in robotics for their high load-carrying capacity and lightweight design. Based on this design, our research group developed the TM-40 tensegrity manipulator, which draws inspiration from the biological musculoskeletal systems shown in Fig.~\ref{fig:TM40 tensegrity manipulator} (b). It has adjustable compliance, which allows the structure to be tuned between softer and stiffer configurations as required. In addition, it possess higher number of actuators than DOFs, which makes it highly redundant.

In general, with rigidity and flexibility, they are also lightweight and have high strength-to-weight ratios, which make them suitable for use in many applications in soft robotics, such as twist manipulation \cite{kobayashi2023large}, precision tasks in confined spaces \cite{10891749} and navigation \cite{doi:10.1089/soro.2022.0048}, etc. Similar to soft continuum manipulators, tensegrity manipulators do not have discrete joints, which makes the shape estimation an open research problem \cite{shah2022tensegrity}. In rigid manipulators, shape or configuration is often considered a trivial task due to the presence of fixed joints, and it can be primarily achieved using conventional sensors such as encoders or potentiometers. In tensegrity structures, the rigid elements (struts) are not connected directly with joints, behaving like floating bodies that are stabilised by the pretension cable elements. As a result, estimating the shape of a tensegrity robot using only the sensors inside its body is an inherently challenging task. 

Theoretically, in actively driven tensegrity robots, if we know the lengths of all cables and strut elements, we can estimate the shape of the structure, as shown by \cite{tong2025tensegrity}  under geometric constraint, the shape can be determined by cable length measurements. However, most tensegrity robots do not actively control all cable lengths; some cables are passive, so their lengths are not directly measurable. This creates an incomplete measurement of the cable lengths, and it is typically compensated for by using additional sensors. For example, Johnson \textit{et al.}\cite{johnson2022sensor} presented a stretchable sensor that is used to calculate the length of the passive cables, and their subsequent work \cite{lu20226n} has demonstrated on a real tensegrity robot. Likewise, Mao \textit{et al.}\cite{10696979} integrated multimodal strain sensors, which replace all traditional cables with these sensors. Booth \textit{et al.}\cite{booth2021surface} proposed strain sensors, which replace this missing information using the strain-based surface sensing.

Many of the studies, as mentioned earlier, are focused on measuring the cable lengths. Another promising approach is to use the Inertial Measurement Unit (IMU) sensors to estimate the shape. Previous research has shown that the IMU sensors have demonstrated considerable promise in reconstructing the posture \cite{santaera2015low}, and kinematic model \cite{sato2024robot } of the rigid robots. In deformable manipulators, previous studies \cite{stella2023soft, pei2025polynomial, cheng2022orientation,martin2022proprioceptive, pei2024imu } have demonstrated the use of IMUs for shape reconstruction. In tensegrity-based robots, significant progress has been made, for instance, Wada \textit{et al.} \cite{10974957} utilised IMUs to define the posture of the tensegrity manipulator based on the tilt of the strut elements. Several other studies \cite{caluwaerts2016state, bezawada2022shape} combined IMU sensing with external sensors to estimate the robot's state and shape. However, to the best of our knowledge, no existing work estimates shape solely based on a single sensor type, such as IMU sensors. This remains a challenge with tensegrity systems, because developing a shape estimation method that depends on a single sensor type, such as IMU sensors, could be generalised to other tensegrity-based robots. This is because these sensors are easy to mount, and we do not need major design modifications to use them. 

Many of these studies, as mentioned above, have been validated on a single-layer tensegrity structure. There is currently no study that demonstrates the scalability of these methods on a full-scale tensegrity manipulator. To address this, we presented a shape estimator based on the cable energies of the tensegrity structure to reconstruct the shape. This work extends prior work\cite{10955232}, which demonstrated the efficacy of this algorithm on a simple single-layer tensegrity structure. Our main contributions can be summarised as follows:

\begin{itemize}
    \item We extend the shape estimator for the complex full-scale tensegrity manipulator based on the proposed approach.

   \item We demonstrate the applicability of the method on a highly redundant five-layer tensegrity manipulator.
   
   \item We show that the shape estimator consistently converges under different initial conditions and estimates the external disturbances.

\end{itemize}

To test our method, we use the TM40 manipulator, which consists of five class-1 tensegrity modules, 20 struts, and 80 tensile members. We calculate the inclination angles of the strut elements with respect to the gravity vector. The method estimates the spatial nodal positions of the structure using energy minimisation. These positions define the overall shape of the manipulator, shown in Fig.~\ref{fig:TM40 tensegrity manipulator} (d), (e).  The results show that the estimated shape stabilises in a minimal energy state, and forms a shape which is approximately close to the actual shape.

This paper is organised as follows: Section II shows the formulation of the shape estimation problem. Section III presents the implementation, results and discussions and finally Section IV concludes and provides directions for future research.

\section{Problem formulation}
\begin{figure}[t]
\centering
    \includegraphics[width=93mm,height=38mm]{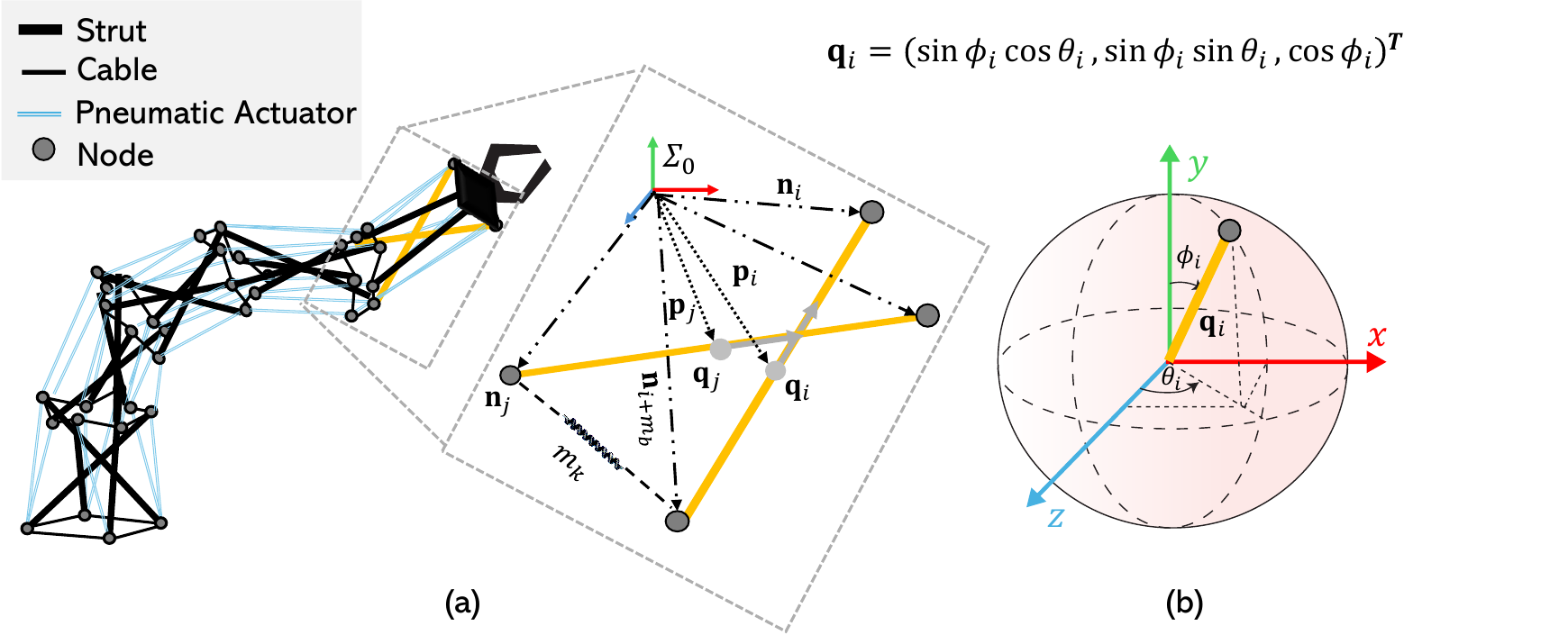}
    \caption{Schematic representation of the tensegrity manipulator. a) Geometric representation of the nodal positions, centre positions, and orientations of the strut elements. b) Transformation of the strut orientation from spherical to Cartesian coordinates. }
    \label{fig:ith_strut_TM40}
    \end{figure}

This section formalises the shape reconstruction problem for the tensegrity manipulator. For a more detailed explanation of this formulation, the reader may refer to \cite{10955232}. Given the inclination angles of the strut elements, the method determines the spatial nodal positions of the structure. For the \textit{i-th} element, shown in Fig.~\ref{fig:ith_strut_TM40} (a),  the nodal positions or end positions of the strut can be written as.
 \begin{equation}
\begin{aligned}
\mathbf{n}_i & = \mathbf{p}_i+ \frac{L_i}{2} \mathbf{q}_i (\phi_i, \theta_i), & 
\mathbf{n}_{i + m_b} & = \mathbf{p}_i - \frac{L_i}{2} \mathbf{q}_i(\phi_i, \theta_i)
\end{aligned}
\label{eq:two_nodes}
\end{equation}
where \( \mathbf{n}_i \in \mathbb{R}^3  \) and  \(\mathbf{n}_{i + m_b}\in \mathbb{R}^3 \) represent the positions of the two nodes connected to the strut \( i \). Here \( i \) ranges from 1 to \( m_b \), with \(m_b \) is the total number of struts in the structure. \( L_i \) is the length of the strut element, \(\mathbf{p}_i\in \mathbb{R}^3 \) is the centre strut position, \(\mathbf{q}_i \in \mathbb{R}^3 \) is a unit vector ($\|\mathbf{q}_i\|$ = 1) parallel to the strut's longitudinal direction, $\phi_i $ and $\theta_i$ are the inclination and yaw angle of the strut element respectively, as shown in Fig.~\ref{fig:ith_strut_TM40} (b).

The whole TM40 arm's shape is fully defined by the 3-dimensional positions of its nodes, which can be written as.
\begin{equation}
\mathbf{n} = \mathbf{A}^T \mathbf{p} + \mathbf{B}^T \mathbf{q} \in \mathbb{R}^{120\times{1}}
\label{eq:nodalvector}
\end{equation}
where,
\[
\mathbf{A}^T = \left[ \begin{array}{c} \mathbf{I}_{3m_b} \\ \hline \mathbf{I}_{3m_b} \end{array} \right] \in \mathbb{R}^{120 \times 60} , 
\mathbf{B}^T = \left[ \begin{array}{c} \frac{\mathbf{L}}{2} \\[1mm]

\hline  \\[-3.5mm] -\frac{\mathbf{L}}{2}  \end{array} \right] \in \mathbb{R}^{120 \times 60} \]
 Where the $\mathbf{p}\in \mathbb{R}^{3m_b}$ and $\mathbf{q}\in \mathbb{R}^{3m_b}$ are the vectors, which consist of all the strut centre spatial positions and the orientations of the strut elements, respectively. $\mathbf{I}_{3m_b}\in \mathbb{R}^{3m_b}$ denotes an identity matrix, and $\mathbf{L}$ is a diagonal matrix defined as \( \operatorname{diag}(L_1 \cdots L_{3m_b}) \).
 \subsection{ Elastic Energy of the Cables }

 The total elastic energy \( e \) stored in all cable elements is the sum of the individual cable energies, $e_k$, and it can be expressed as.
 \begin{equation}
\label{eq:deqn_ex2}
e = \sum_{k=1}^{m_s}   e_k,e_k= \begin{cases}\frac{1}{2} K_k\left(m_k-b_k\right)^2, & m_k>b_k \\ 0, & m_k \leq b_k\end{cases}
\end{equation}
where \( K_k \), $m_k=\left\|\mathbf{n}_j-\mathbf{n}_{i+m_b}\right\|$, and \(b_k\) correspond to the stiffness, current length, and the natural length of the \textit{k-th} cable element, respectively, as shown in Fig.~\ref{fig:ith_strut_TM40} (a). Here, \(m_s\) is the total number of cables in the structure.
\begin{figure*}[tb]
 \centering
    \includegraphics[width=180mm,height=48mm]{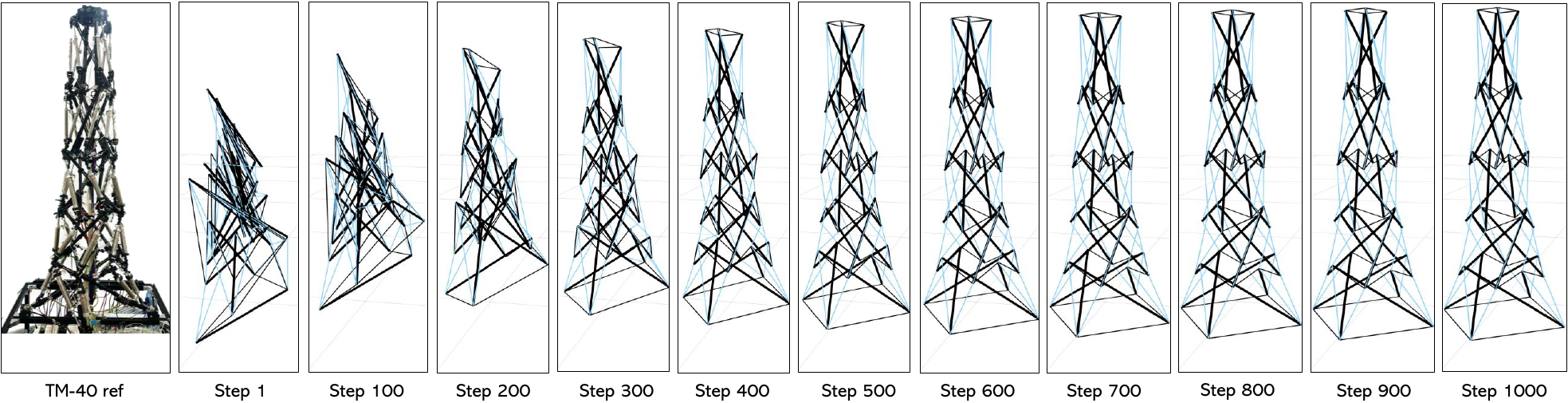}
     \caption{The figure shows the visualisation of the optimisation process in RViz2. Given the inclination angles of the strut elements, the method estimates the spatial nodal positions of the structure by minimising the energy function.}
     \label{fig:states}

\end{figure*}

In general form, the total elastic potential energy stored in all the cable elements of the structure can be written as:
\begin{equation}
\begin{aligned}
e = \frac{1}{2}  \mathbf {(m_s - b_s)}^T \mathbf {K} \mathbf {(m_s-b_s)}
\end{aligned}
\label{eq:energy_in_summation}
\end{equation}

Equation (\ref{eq:energy_in_summation}) represents the quadratic form of the elastic energy, where \( \mathbf{K} = \operatorname{diag}(K_1, \dots, K_{3m_s}) \) represents the stiffness matrix, $\mathbf {b_s}$ is a vector of natural lengths, and $\mathbf {m_s}$ represents the current lengths of all the cable elements and is defined as :
\begin{equation}
\begin{aligned}
& \mathbf {m_s} = \mathbf {C_s} \left( \mathbf {A} ^T \mathbf {p} + \mathbf {B} ^T \mathbf {q}  \right)
\end{aligned}
\label{eq:ms}
\end{equation}

where $\mathbf{C_s} \in \mathbb{R}^{3 m_s \times 3 n}$ is a connectivity matrix for the cable elements. It consists of values 1, 0, and -1, with each row indicating how the cable elements in the structure are connected to nodes. Here, $n$ is the total number of nodes in the structure.

The equation \ref{eq:energy_in_summation} can also be written as :
\begin{equation}
\begin{aligned}
e & = \frac{1}{2} \left( \mathbf{C_s} \mathbf{A}^T \mathbf{p}+ \mathbf{C_s} \mathbf{B}^T \mathbf{q(\boldsymbol{\phi},\boldsymbol{\theta})} - \mathbf{b_s} \right)^T \mathbf{K} ( \mathbf{C_s} \mathbf{A}^T \mathbf{p} + \\
  &\quad  \mathbf{C_s} \mathbf{B}^T   \mathbf{q(\boldsymbol{\phi},\boldsymbol{\theta})} - \mathbf{b_s} )
\end{aligned}
\label{eq:e_elongation}
\end{equation}

If we ignore the rotation around the longitudinal axis of the struts, then the orientation vectors $\mathbf{q}$ depends only on the inclination angles $\mathbf{\boldsymbol{\phi}}$ ( with respect to the gravity vector) and yaw angles $\mathbf{\boldsymbol{\theta}}$ (around the gravity vector) of the struts.

To obtain stable inclination angles $\mathbf{\boldsymbol{\phi}}$, we use 6-axis gyroscope and accelerometer sensors. The yaw angles $\mathbf{\boldsymbol{\theta}}$ are traditionally determined using magnetometer measurements with gyro integration.  However, yaw estimation is challenging due to magnetic interference in magnetometers and drift over time in gyroscopes. This issue is particularly significant in tensegrity robots because of their geometry. The sensors attached to the struts are in close proximity, which causes magnetic interference. As a result, we rely only on the inclination angles $\mathbf{\boldsymbol{\phi}}$ of the strut elements, while the yaw angles $\mathbf{\boldsymbol{\theta}}$  are treated as unknown parameters in the optimisation problem.
 \subsection{Shape Reconstruction Objective Function}

  A tensegrity structure consists of cables and strut elements, and the points of intersection between these elements are known as “nodes". The forces at these nodes are balanced, which corresponds to the structure being in a minimum energy state \cite{skelton2009tensegrity}, thus the problem of determining unknown parameters $\mathbf{p}$ and $\mathbf{\boldsymbol{\theta}}$ in the structure can be reduced to an energy-minimization problem, and it can be written as.

\begin{equation}
\mathbf{\tilde{p}} , \boldsymbol{\tilde{\theta}}\ = \underset{\boldsymbol{p}, \boldsymbol{\theta}}{\operatorname{argmin}} \, e
\label{eq:arg_min}
\end{equation}
We aim to minimise the energy function $e$ to estimate the nodal positions $\mathbf{n}$ of the structure. The Fig.~\ref{fig:states} shows a visualisation of the optimisation process, while satisfying the optimal conditions.

The optimal conditions are given by the vanishing of the analytically derived gradients.

\begin{equation}
\begin{gathered}
\frac{\partial e}{\partial \mathbf{p}} =\frac{ \partial \mathbf {m_s}}{\partial \mathbf{p}} \mathbf{K}\left[\left(\mathbf{m_s}-\mathbf{b_s}\right) \odot \delta\right] \in \mathbb{R}^{1 \times 3m_b} \\
\frac{\partial e}{\partial \boldsymbol{\theta}}= (\mathbf{K}\left[\left(\mathbf{m_s}-\mathbf{b_s}\right) \odot \delta\right])^T \mathbf{C}_s \mathbf{B}^T \frac{\partial \mathbf{q}}{\partial \boldsymbol{\theta} } \in \mathbb{R}^{1 \times m_b} 
\label{eq:gradients_with_length}
\end{gathered}
\end{equation}

where $\odot$ is a Hadamard product, $\delta \in \{0,1\}^{m_s}$ denotes a mask vector, and its components can be written as

\begin{equation}
\delta_k= \begin{cases}1, & \text { if } m_k-b_k>0 \quad \text { (cable in tension) } \\ 0, & \text { otherwise (slacked cable) }\end{cases}
\end{equation}

If a cable is in a slacked condition $ m_k< b_k$, the force in the cable becomes negative, and the gradient will push the cable apart to reduce the energy instead of pulling it together. 

For simplification,  we assume the natural lengths $\mathbf {b_s}$ of the cable elements are zero. Therefore, the gradient expressions in the equation (\ref{eq:gradients_with_length}) reduce to:
\begin{equation}
\frac{\partial e}{\partial \mathbf{p}} = \frac{\partial \mathbf{m_s}}{\partial \mathbf{p}}\left(\mathbf{K} \mathbf{m_s}\right) , \quad \frac{\partial e}{\partial \boldsymbol{\theta}} = \left(\mathbf {K} \mathbf {m_s} \right)^T  \mathbf {C_s} \mathbf {B}^T  \frac{\partial \mathbf {q}}{\partial \boldsymbol{\boldsymbol{\boldsymbol{\theta}}}}
\label{eq:gradient_p_theta}
\end{equation}

                                            As mentioned earlier, the complexity of obtaining yaw angles $\mathbf{\boldsymbol{\theta}}$ using magnetometer and gyro sensors. So we incorporated the $\mathbf{\boldsymbol{\theta}}$ into the optimisation problem as an unknown variable, and its corresponding gradient is derived in equation  \ref{eq:gradient_p_theta}. In the next step, we compute the Jacobian \( \frac{\partial \boldsymbol {q}}{\partial \boldsymbol {\theta}} \in \mathbb{R}^{3 m_b \times m_b} \) of this gradient by transforming the orientation from spherical to Cartesian coordinates, shown in Fig.~\ref{fig:ith_strut_TM40} (b), the transformation can be written as.
 \begin{equation}
\begin{aligned}
q_{ix} &= \sin \phi_i \cos \theta_i, & 
q_{iy} &= \sin \phi_i \sin \theta_i, & 
q_{iz} &= \cos \phi_i
\end{aligned}
\end{equation}
\begin{figure*}[t]
\centering
    \includegraphics[width=\linewidth]{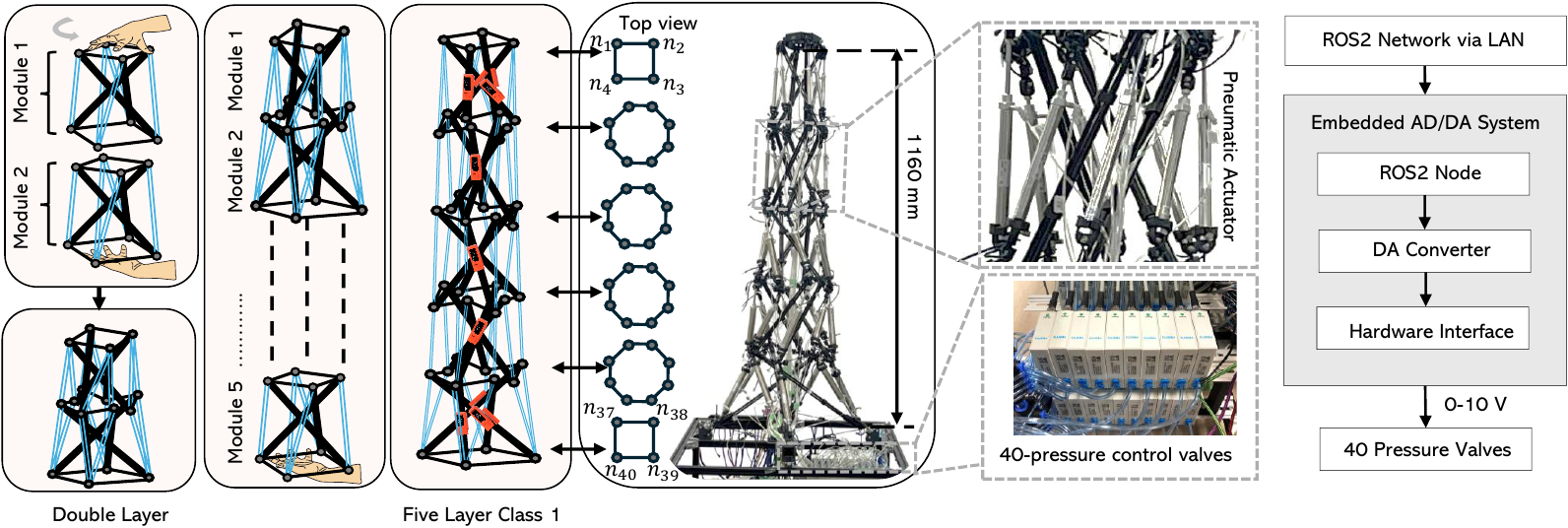}
    \caption{\textbf{Five-Layer tensegrity manipulator:} Each module is stacked with alternating twist direction to form a five-layer tensegrity manipulator. The structure consists of 80 active and passive tension elements and 20 rigid struts. Each pneumatic cylinder is controlled independently, and the system has a ROS2-compatible embedded AD/DA module to actuate the 40 pneumatic cylinders. }
    \label{fig:stacked5layers}
    \end{figure*}
The derivatives of each component of \(\mathbf{q}_i\) with respect to the yaw angle can be written as follows.
\begin{equation}
\begin{aligned}
\frac{\partial q_{ix}}{\partial \theta_i} &= -\sin \phi_i \sin \theta_i, & 
\frac{\partial q_{iy}}{\partial \theta_i} &= \sin \phi_i \cos \theta_i, & 
\frac{\partial q_{iz}}{\partial \theta_i} &= 0
\end{aligned}
\end{equation}

Finally, we solve the optimisation problem numerically using the first-order optimiser, called the gradient descent (GD) method. The unknown parameters $\mathbf{p}$ and $\mathbf{\boldsymbol{\theta}}$ are updated as follows.

\begin{equation}
\mathbf{p}_{k+1}=\mathbf{p}_k-\alpha \frac{\partial e}{\partial \mathbf{p}},     \boldsymbol {\theta}_{k+1}=\mathbf{\boldsymbol{\theta}}_k-\beta \frac{\partial e}{\partial \boldsymbol {\theta}}
\end{equation}

where, $\alpha$ and $\beta$ are the learning rates. Furthermore, when the gradients in equation \ref{eq:gradient_p_theta} satisfy the first-order optimal conditions  \( 
\left\|\frac{\partial e}{\partial \mathbf{p}}\right\| \approx 0
\) and  \( 
\left\|\frac{\partial e}{\partial \boldsymbol {\theta}}\right\| \approx 0
\), which indicates the structure converges to the minimal energy state.
\section{ Experimental Setup}
This section provides a detailed description of the Tensegrity manipulator-TM40 used in the experiments. We then give the details on the sensor module for measuring the inclination angles $\boldsymbol{\phi}$ of the struts.

\subsection{ Description of Tensegrity Manipulator-TM40}
We conducted our experiments on a highly redundant 20-DOF tensegrity manipulator. It consists of 20 struts made of CFRP (Carbon Fibre Reinforced Plastic) pipes, 40 nodes, and 80 tensile members, of which 40 tensile members are passive cables made from Kevlar braided cords and the other 40 tensile members are actively controlled pneumatically by changing the pressure inside the cylinder using 40 pressure control valves (VEAB, FESTO). Each valve controls the pressure inside one cylinder independently. The operating range of pressure values is 0.1 - 0.6 MPa, which corresponds to the analog voltage values 0-10 V. The embedded AD/DA system, which is ROS2-compatible, provides target pressure values.

Each module of the continuum consists of 4 rigid struts, and the modules are stacked sequentially to form a five-layer continuum manipulator, as shown in Fig.~\ref{fig:stacked5layers}. The strut lengths vary across the manipulator: the upper four modules have struts with an end-to-end length of approximately 330 mm, while the bottom modules' (Module 5) struts measure about 390 mm. Each strut has an outer diameter of 5 mm.

At the top and bottom, each module contains the square patterns, and when we connect these patterns, they form an octagon pattern. In total, the structure has two square and four octagonal closed loop paths formed by passive cables under high tension, which provides stability to the structure. In addition, 40 active pneumatic cylinder cables are placed vertically between these closed paths, with each module having 8 pneumatic cylinders. These pneumatic cylinders can independently extend or contract, and can pull these closed-loop paths closer together, thereby inducing bending of the manipulator. The modules are stacked in two directions. For instance, the Module 1 stacks on the Module 2 with a twist in a counterclockwise direction, and a Module 3 twist in a clockwise direction, this balances out the twisting forces and makes the whole system more stable. 

 \begin{figure}[h ]
\centering
    \includegraphics[width=90mm,height=30mm]{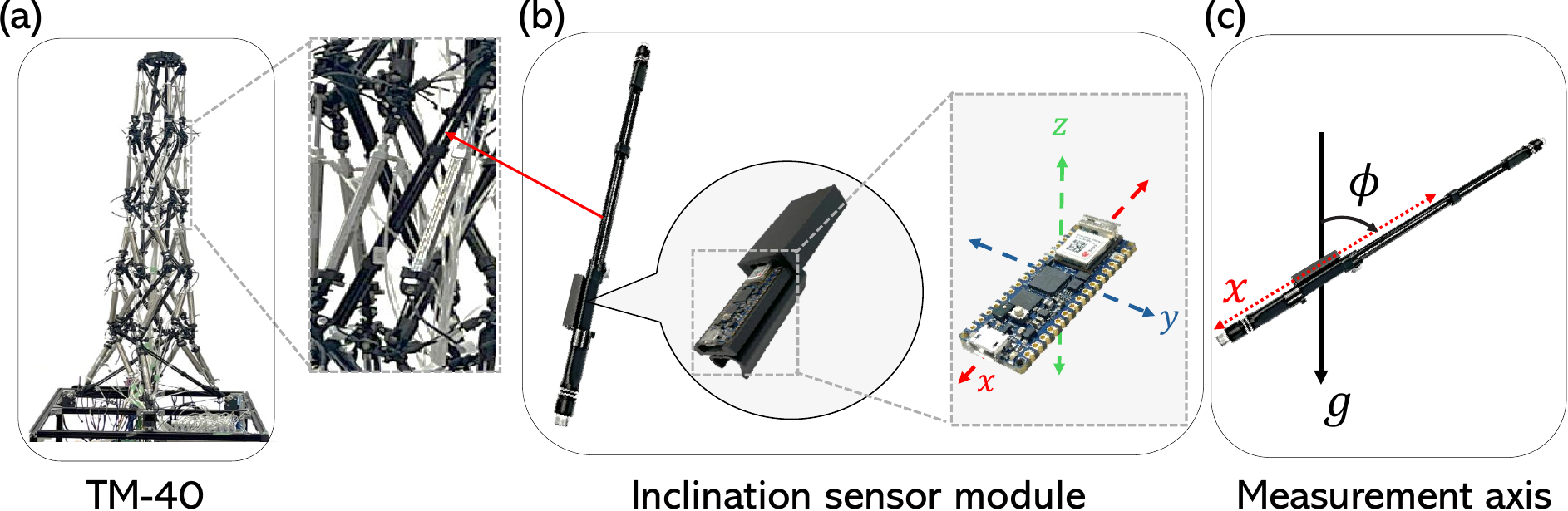}
    \caption{\textbf{Sensor module for inclination angles:} a) The manipulator consists of 20 struts. b) Each is equipped with a 6-axis IMU sensor. c) The inclination angles are measured from the longitudinal axis direction of each strut element and the gravitational vector.}
    \label{fig:inclination_module}
    \end{figure}

\subsection{Inclination Angle Sensor Module}
The inclination angles $\boldsymbol{\phi}$ are measured from the strut longitudinal direction with the gravity vector $g$, as shown in Fig.~\ref{fig:inclination_module} (c). We used an Arduino Nano RP2040 equipped with an onboard IMU sensor (LSM6DSOX). The sensor measures 3-axis accelerometers, $(a_x, a_y, a_z)$ and 3-axis gyroscope $(w_x, w_y, w_z)$. Under quasi-static conditions, the inclination angle of the strut can be computed as:
\begin{eqnarray}
{\phi}_q = \arccos \frac{\boldsymbol{a} \cdot \boldsymbol{a_g}}{|\boldsymbol{a}||\boldsymbol{a_g}|}
= \arccos \frac{a_x}{|\boldsymbol{a}|}
\end{eqnarray}
where $\boldsymbol{a} = (a_x, a_y, a_z)$ and  $\boldsymbol{a_g} = (g, 0, 0)$. In dynamic conditions, the inclination angles based on the accelerometer sensor lead to errors, so we use the angular velocity around the axis perpendicular to the plane of inclination.
\begin{equation}
    \omega = \frac{a_y \cdot w_z - a_z \cdot w_y}{\sqrt{a_y^2 + a_z^2}}
\end{equation}

Here, $w_y$, $w_z$ represent the angular velocity readings around the $y$ and $z$ axes, respectively. To ensure stable inclination angle measurements, we fused the accelerometer and gyroscope sensor readings using a complementary filter, and its update rule can be written as:
\begin{eqnarray}
    \phi_{t+1} = \frac{T_c}{T_c+T_s} (\phi_{t}+T_s \omega_t) + (1-\frac{T_c}{T_c+T_s}) ({\phi}_q)_t
\end{eqnarray}
where $T_c$ is a time constant and $T_s$ is sampling period. In our experiments, we set $T_c$ and $T_s$ to 200 \textit{ms} and 12.5 \textit{ms}, respectively.
\begin{figure}[hbt]
\centering
    \includegraphics[width=80mm,height=47mm]{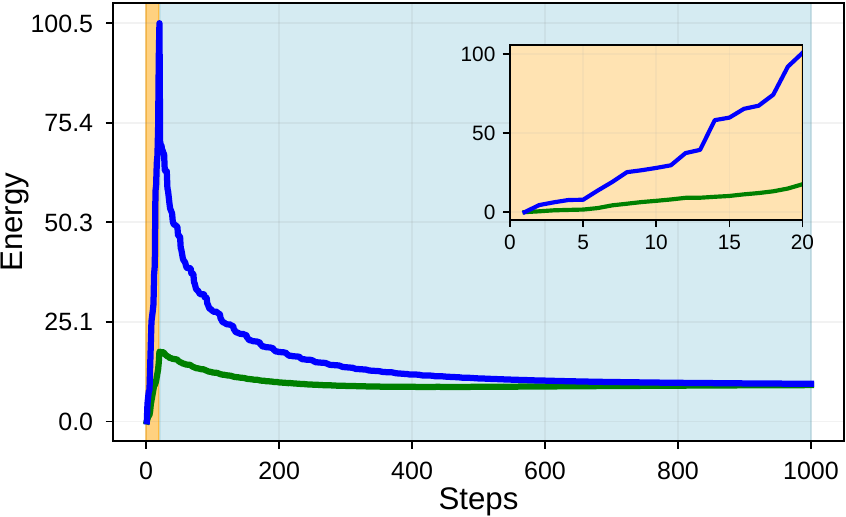}
    \caption{The plot shows the energy of the structure in the collapsed (green) and expanded (blue) states. The yellow-shaded region represents the cumulative energy of all cable elements for the first 20 steps of optimisation, while the light blue-shaded region shows the decrement in total energy over the optimisation steps.}
    \label{fig:energy_plot_expn_cont}
    \end{figure}
To transmit the data to the Robot Operating System 2 (ROS2) network, we used the microros agent, which bridges between the ROS2 stack and the micro-ROS client node, running in the microcontroller. The communication between the micro ROS node client and the agent occurs over a serial connection at 80 Hz.

\begin{figure*}[tb]
\centering

\begin{minipage}{0.32\linewidth}
  \centering
  \includegraphics[width=\linewidth]{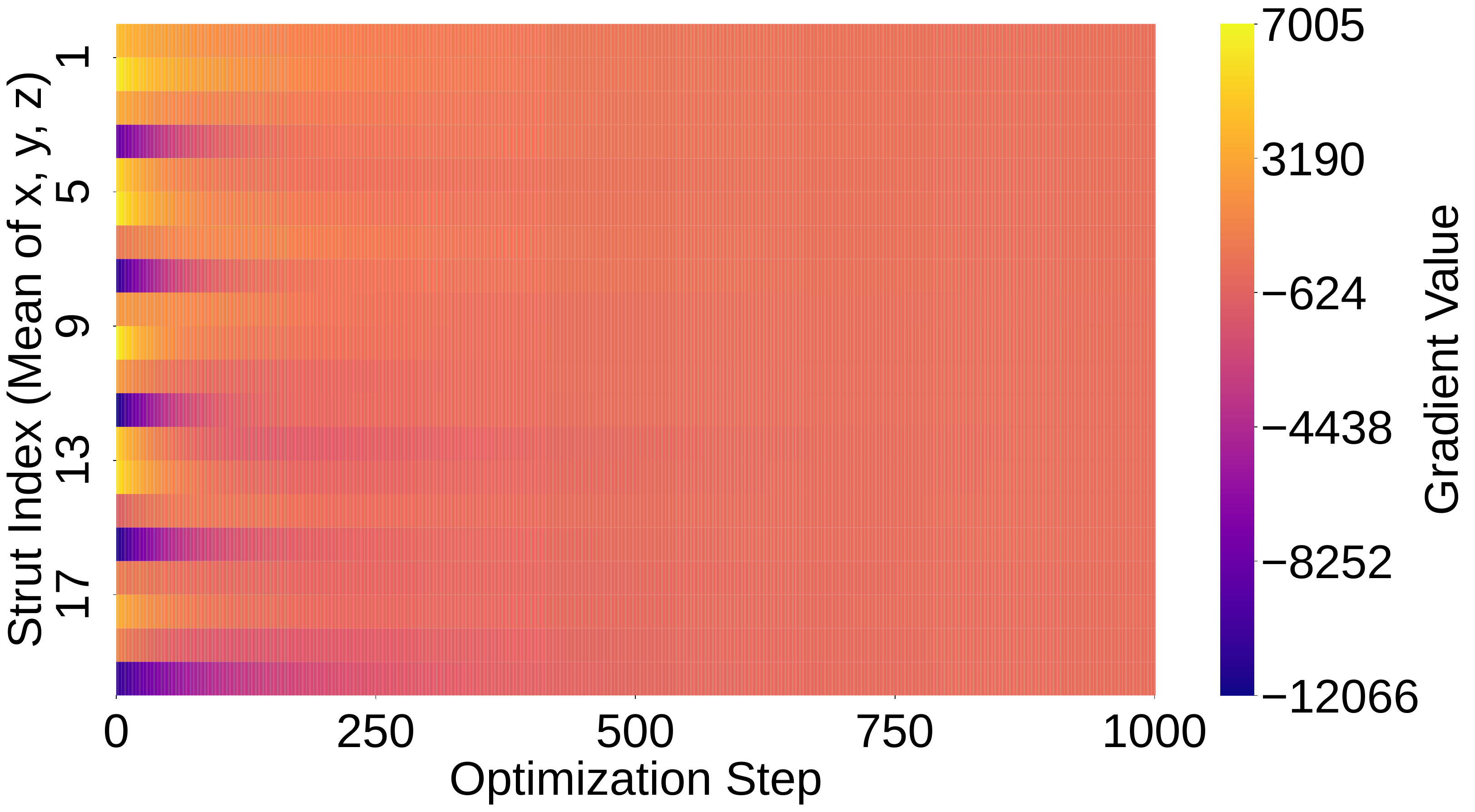}
  \label{fig:mean_gradients_per_strut_Exp}
\end{minipage}
\hfill
\begin{minipage}{0.32\linewidth}
  \centering
  \includegraphics[width=\linewidth]{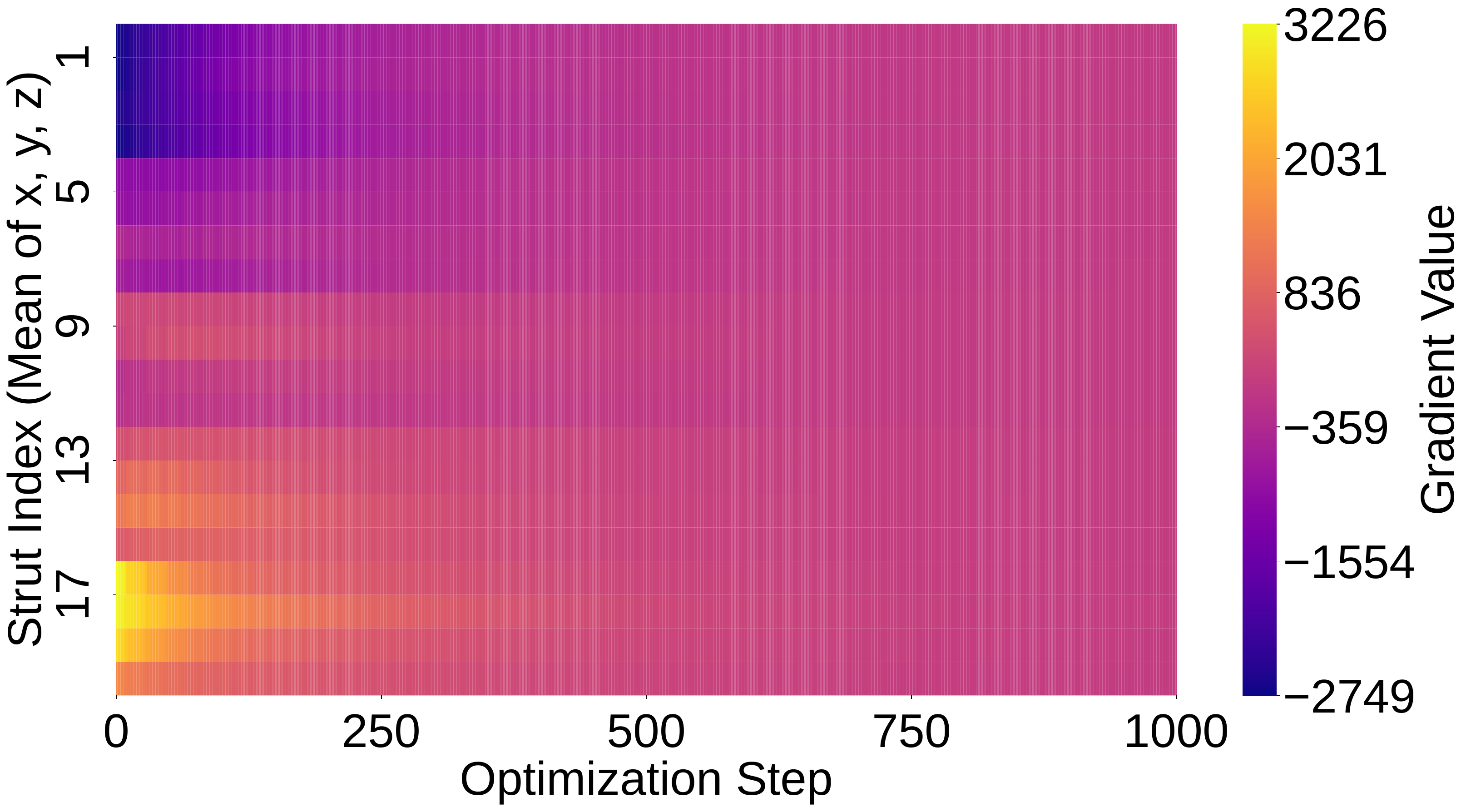}
  \label{fig:mean_gradients_per_strut_Cont}
\end{minipage}
\hfill
\begin{minipage}{0.32\linewidth}
  \centering
  \includegraphics[width=\linewidth]{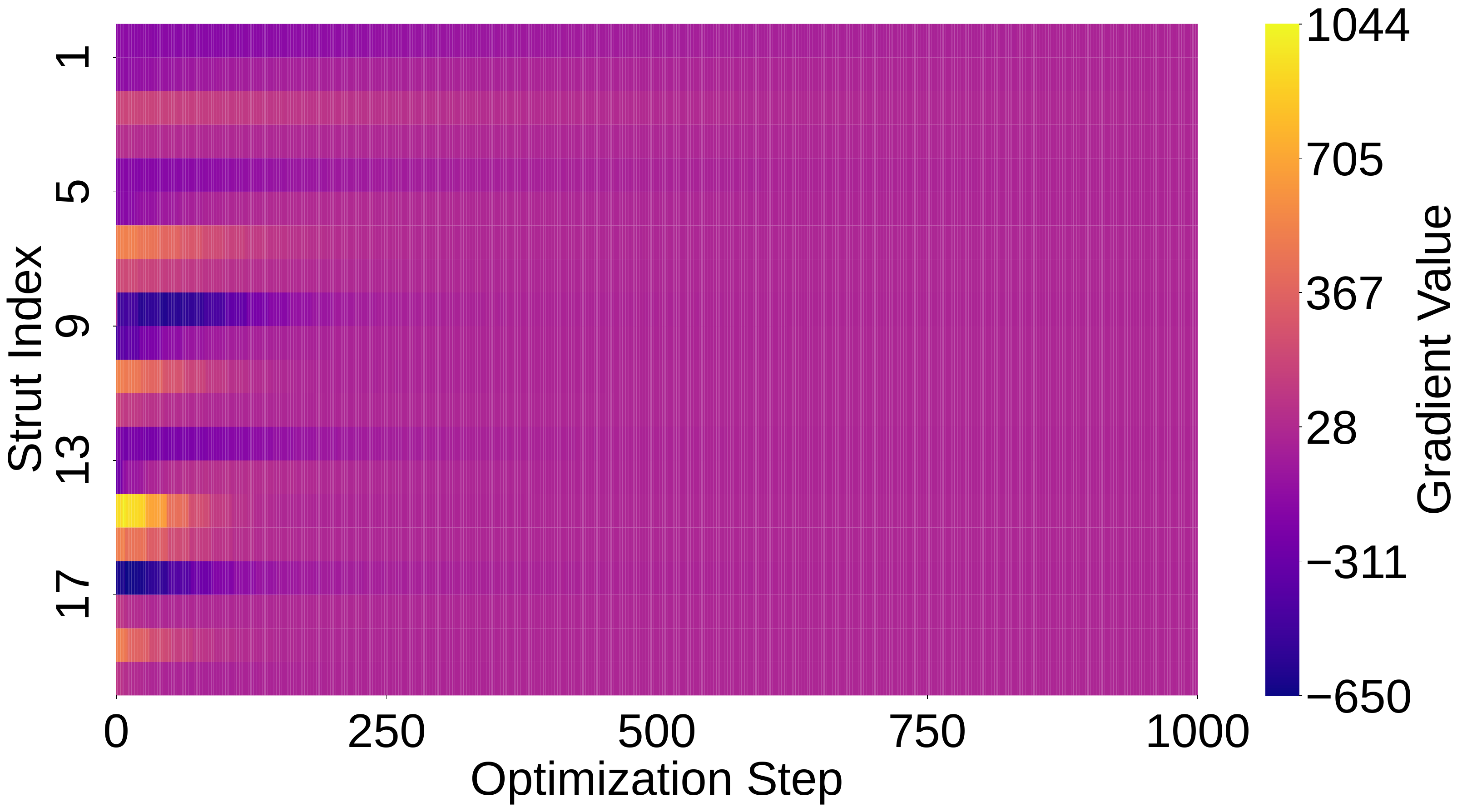}
  \label{fig:yaw_gradients_heatmap}
\end{minipage}

\caption{The heatmaps in the left and middle panels show the gradients of the energy function with respect to the mean of $x$, $y$, and $z$ centre positions during the collapsed and expanded states, respectively. The right panel shows gradients with respect to the yaw angles of the struts. Each row corresponds to a strut element. In all the heatmaps, the gradients gradually vanish as the optimisation steps increase, which indicates convergence.}
\label{fig:both_heatmap}
\end{figure*}
  \begin{figure*}[t]
\centering
\includegraphics[width=180mm,height=35mm]{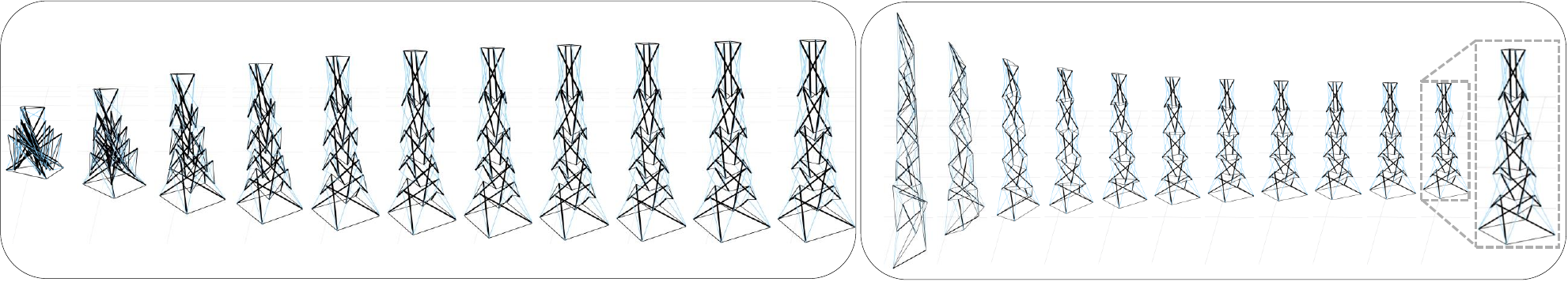}

\caption{The left and right sides show estimations starting from collapsed and expanded states, respectively. In both cases, the estimated shape gradually converges toward the actual shape as the optimisation steps increase. By the 1000th step, both cases reach a minimal energy state.}
\label{fig:collapsed_expanded_snaps}

\end{figure*}

\section{ Experimental results and discussion}
In this section, we present the experimental studies along with their discussions. First, the proposed method is evaluated under static conditions with two extreme cases. Second, to test the robustness, we evaluated the shape estimator using 10 different initial conditions. Finally, we then externally induced manual deformations to the manipulator to evaluate its behaviour.

\subsection{ System Configuration}
The proposed method of shape estimation was implemented in Python 3.10, and the experiments were performed on a desktop computer with an Intel Core i9-10885H processor and Ubuntu 22.04 LTS as the operating system, with ROS 2 Humble as the software environment. In addition, we used the ROS's visualisation tool (RViz2) to display the results as markers. Furthermore, the open and non-continuous lattice structure of the tensegrity manipulator makes it difficult to place markers on each strut element, which makes it difficult to obtain the ground truth measurements for qualitative assessment of the whole body estimated shape or the nodal positions. Therefore, we performed a quantitative evaluation of the shape estimator by comparing the estimated shape with the actual shape of the physical tensegrity manipulator. 

\subsection{Static Case}
In the static experiment, we pressurised all the 40 active cables symmetrically with the pressure values \(\mathbf{p_c} \in \mathbb{R}^{40}\), each set to 0.6 MPa, and for each corresponding pressure value we manually tuned the stiffness matrix $\mathbf{K}$. The remaining 40 passive cable elements we set relatively higher stiffness coefficients. The coefficients were set in a ratio of 140:27 (passive:active cables), tuned so that the shape estimator achieves a solution without oscillatory behaviour and maintains a desirable convergence speed. We run the optimisation for 1000 steps. At each step, the energies of the cables attached to the current strut are calculated and updated using the GD method. After each step, the centre position and yaw angle are recalculated. Every 20 steps, all 20 IMU measurements attached to the struts of the TM-40 manipulator are sequentially obtained, which completes one full update of the structure.

\subsubsection{Collapsed state}
In the collapsed state, we initialised the unknown parameter $\mathbf{p}$ such that the estimated shape starts from a collapsed state at the first step of optimisation. The method computes the energy $e$ of the structure over the first 20 optimisation steps, during which the estimated shape reaches a peak energy of 17.61 units. Beyond this step of optimisation, the energy $e$ decreases monotonically, as shown in Fig.~\ref{fig:energy_plot_expn_cont}. As the number of optimisation steps increases, the estimated shape expands. Mathematically, gradients $\frac{\partial e}{\partial \mathbf{p}}$ and $\frac{\partial e}{\partial \boldsymbol {\theta}}$
in (\ref{eq:gradient_p_theta}) gradually diminish. As shown in Fig.~\ref{fig:both_heatmap} (left, right), the gradients initially have a large magnitude, and by the 1000th step, they approximately have zero value, which is also shown by the fading of colour intensity. Figure~\ref {fig:collapsed_expanded_snaps} (left side) shows that the estimated shape converges to a minimal energy state, with the energy value fluctuating around 9.15 units, which approximately corresponds with the shape of the actual structure. 

\subsubsection{Expanded state}
In the second case, we set different initial values, and the structure began from the expanded state, shown in Fig.~\ref{fig:energy_plot_expn_cont}. The computed peak energy in this case reaches 100.54 units,  which is much higher than in the first case. This is because the initially estimated nodal positions are more widely separated, thus storing more energy. Consequently, as the optimisation steps increase, the estimated shape converges to a neighbourhood of the minimal energy state, with an energy value of approximately 9.48 units, shown in Fig.~\ref {fig:collapsed_expanded_snaps} (right side). Table~\ref{tab:manipulator_length} shows the estimated length of the manipulator at the 1000th step of optimisation.  We calculate the estimated length of the manipulator using. 
\[
\begin{aligned}
d &= \left\| \frac{1}{4} \Bigg( \sum_{k=1}^{4} \mathbf{r}_{n_k} - \sum_{k=37}^{40} \mathbf{r}_{n_k} \Bigg) \right\|
\end{aligned}
\]
where, $\mathbf{r}_{n_k}$ is the position vector, $k = 1\dots4$ are the top nodes of the Module 1 and  $k = 37\dots40$ are the bottom nodes of Module 5. The estimated length calculated is 1135.50 mm, with an absolute error of 24.5 mm relative to the actual length. The potential sources of error may arise from inaccuracies in the IMU inclination angles and from the need for further tuning of the stiffness matrix.
\begin{figure}[tb]
\centering
    \includegraphics[width=80mm,height=47mm]{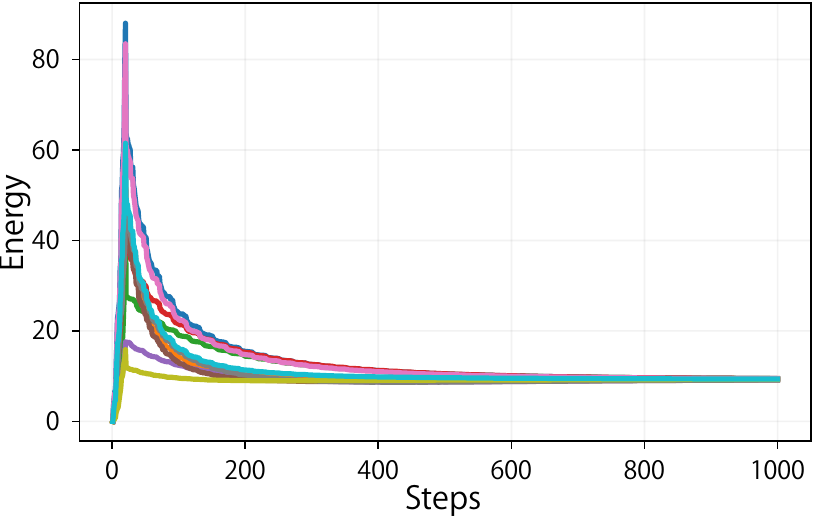}
    \caption{The estimated shape initialised from random unknown parameters. The figure shows that the optimisation problem consistently converges to the same optimal solution across all 10 experiments.}
    \label{fig:10_experiments}
    \end{figure}
\begin{table}[tb]
\caption{Estimated manipulator length in mm.}
\label{tab:manipulator_length}
\centering
\setlength{\tabcolsep}{1.5pt} % Adjust column separation for compactness
\renewcommand{\arraystretch}{1.05} % Adjust row spacing for readability
\resizebox{\columnwidth}{!}{
\begin{tabular}{lcccccc}
\hline
\textbf{Node Group} & \textbf{Avg. x} & \textbf{Avg. y } & \textbf{Avg. z } & $d$ & \textbf{Error} & \textbf{\% Error} \\ \hline
Top ($\mathbf{n}_1$--$\mathbf{n}_4$) & 4.82 & 4.24 & 655.82 & 1135.5 & 2.11 & 24.5 \\
Base ($\mathbf{n}_{37}$--$\mathbf{n}_{40}$) & -4.26 & -5.70 & -479.59 & -- & -- & -- \\ \hline
\end{tabular}
}
\end{table}
    
To test the robustness, we randomly initialise both the unknown parameters $\mathbf{p}$ and $\boldsymbol{\theta}$ in 10 different cases. As shown in Fig.~\ref{fig:10_experiments}, the optimisation problem converges approximately to the same optimal solution across all the experiments, with an average converged energy value of 9.32 units.  We also computed the running time, on average, the algorithm required 7.3 seconds to reach an optimal solution from the first step of optimisation, satisfying the conditions  \( 
\left\|\frac{\partial e}{\partial \mathbf{p}}\right\| \approx 0
\) and  \( 
\left\|\frac{\partial e}{\partial \boldsymbol {\theta}}\right\| \approx 0
\). After convergence, single step takes $\sim$ 7.36 \textit{ms}, and 20 steps of optimisation requires $\sim$ 147.2 \textit{ms}. The reported times represent average computation times and do not include system latencies. Furthermore, the IMU measurements could be parallelized, which could significantly improve real-time performance.

In addition, as mentioned earlier, quantitative evaluation of the nodal positions is not feasible for our current experimental setup due to occlusions. However, when evaluated on a simple single-layer tensegrity structure, our previous work \cite{10955232} reported that the same algorithm achieved a mean absolute error (MAE) of 3.76 mm for the centre strut positions and 20.78 mm for the nodal positions when using an optimised stiffness matrix $\mathbf{K}$. In this work, we focus primarily on qualitative assessment.
\begin{figure*}[!t]
\centering
\includegraphics[width=\linewidth]{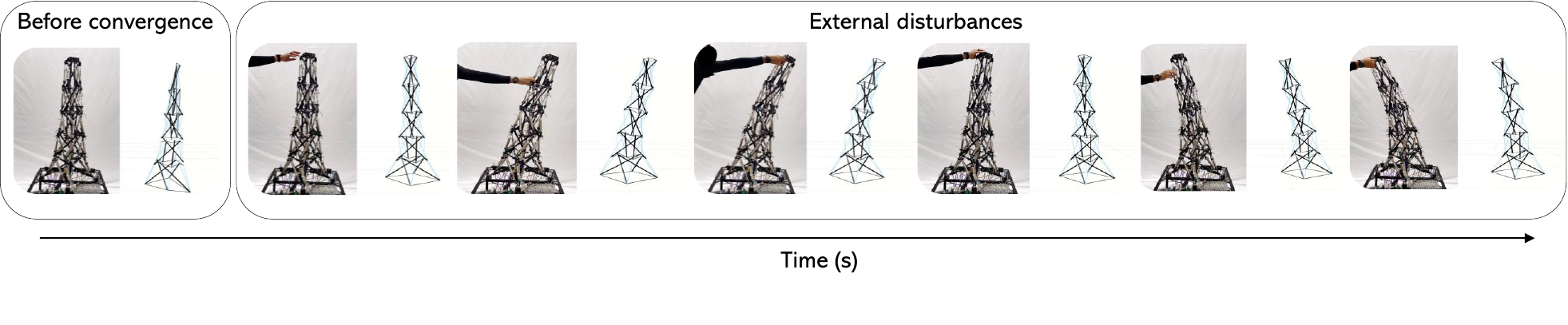}
\vspace{-1.0cm}

\caption
  {The left panel shows the estimated shape of the tensegrity manipulator before convergence. The right panel shows the converged shape, and then the manual external perturbations are applied at the tip of the arm, which induce bending to the left and right. The shape estimator approximately estimated these deformations, and the corresponding shapes of the TM-40 are visualised in RViz2. }
  \label{fig:TM40_snaps}
\end{figure*}
\subsection{External Deformation}
In the external deformation experiment, we further tuned the stiffness matrix $\mathbf{K}$ and initialised the parameters $\mathbf{p}$ and $\boldsymbol{\theta}$ with random values. After around 500 optimisation steps, the estimated shape converges to the actual shape of the tensegrity manipulator. Subsequently, we apply manual external perturbations at the tip of the arm. As shown in the Fig.~\ref{fig:TM40_snaps},  the shape estimation method captures the resulting deformations.

However, in the current experiments, we observed that the estimated shape gets worse when the robot bends more deeply. We believe this is a limitation of not considering the natural lengths of the cable elements. Incorporating the natural lengths in the optimisation problem as constraints is expected to improve the estimation results.  Furthermore, we manually tuned the stiffness matrix $\mathbf{K}$ coefficients of both active and passive cable elements for a particular posture. In practice, however, the stiffness of the active cable elements varies with the applied pneumatic pressure. To address this, we plan to integrate a stiffness model that maps the stiffness values of the active cables to the corresponding pressure values. As a result, the shape of the manipulator can be estimated more accurately during deeper bending and active motion.

\section{Conclusion}
We present a novel proprioceptive shape estimator for a tensegrity manipulator based on the inclination angles of the strut elements. We demonstrated through experiments that our proposed method can estimate the shape of a full-scale five-layer tensegrity manipulator using off-the-shelf IMU sensors, without relying on any external sensing. We believe our energy-based shape estimator can be generalised to other tensegrity-based manipulators with minimal hardware modifications.

In future work, we aim to include the natural length and incorporate the stiffness model for the cable elements to evaluate the deep and active motion bending of the tensegrity manipulator. In addition, we also aim to parallelise the sensor measurements to improve the real-time performance.

\section*{Acknowledgments}
We thank Yuhei Yoshimitsu and Kazuki Wada for developing the inclination sensor module for the tensegrity manipulator. This work was supported by JSPS KAKENHI Grant Number 25H02619.

\bibliographystyle{IEEEtran}
\bibliography{references} 

\end{document}